# WHAT IS AN OPTIMAL DIAGNOSIS?

David Poole* and Gregory M. Provan[†]
Department of Computer Science
University of British Columbia
Vancouver, BC
Canada V6T 1W5

## Abstract

Within diagnostic reasoning there have been a number of proposed definitions of a diagnosis, and of an optimal or most likely diagnosis. These include most probable posterior hypothesis, most probable interpretation, most probable covering hypothesis, etc. Most of these approaches assume that the most likely diagnosis *must* be computed, and that a definition of what should be computed can be made *a priori*, independent of what the diagnosis is used for. We argue that the diagnostic problem, as currently posed, is incomplete: it does not consider how the diagnosis is to be used, or the utility associated with the treatment of the abnormalities. In this paper we analyse several well-known definitions of diagnosis, showing that the different definitions of optimal diagnosis have different qualitative meanings, even given the same input data. We argue that the most appropriate definition of (optimal) diagnosis needs to take into account the utility of outcomes and what the diagnosis is used for.

## 1 INTRODUCTION

Within diagnostic reasoning there have been a number of proposals of what constitutes a diagnosis, and so presumably, what constitutes an optimal or most likely diagnosis. These include most probable posterior hypothesis [Pearl, 1986], most probable interpretation [Pearl, 1987], most probable provable hypothesis [Reiter, 1987, de Kleer and Williams, 1987, de Kleer *et al.*, 1990], most probable covering hypothesis [Reggia *et al.*, 1985, Peng and Reggia, 1987a, Peng and Reggia, 1987b]. Unlike earlier logic-based diagnoses that consider what can be proven about a faulty device [Genesereth, 1984], these papers have considered that the question "what is a diagnosis?" is important to answer. The intuition is that it is important to characterise the set of "logical possibilities" for a diagnosis, presumably to be able to compare them. Most of these approaches assume that the most likely diagnosis *must* be computed, and that a definition of the what should be computed can be made *a priori*, independently of what the diagnosis is used for.

*This author was supported by NSERC grant OGPOO44121.

[†]The author completed this research with the support of NSERC grant A9281 to A.K. Mackworth.

Once the sets of hypotheses considered as diagnoses are determined, one of the ways we may want to compare competing diagnoses to give us the most likely diagnosis is by using probability [de Kleer and Williams, 1987, Neufeld and Poole, 1987, de Kleer and Williams, 1989, Pearl, 1987]. In computing the probability of $A$ given $B$, $p(A|B)$, Bayesian analysis tells us that the $B$ should be all of the available evidence [Kyburg, 1988, Pearl, 1988], but does not give us any hints as to what $A$ should be. This paper asks the question of what combination of hypotheses $A$ should be in order to be most useful.

In this paper we study six approaches to diagnostic reasoning, and their associated notions of optimality based on probability theory (or another uncertainty calculus). Each approach considers a conjunction of hypotheses as a most likely diagnosis. We call the six approaches:

1. most likely single-fault hypothesis;
2. most likely posterior hypothesis;
3. most likely interpretation;
4. probability of provability;
5. covering explanations; and
6. utility-based explanation.

We contrast the first five approaches to diagnostic reasoning with a classic utility-based approach to diagnostic reasoning [Ledley and Lusted, 1959].

In analysing these proposals, we show that the different definitions of optimal diagnosis have different qualitative (and quantitative) results, even given the same input data. Moreover, we argue that the diagnostic problem, as currently posed, is incomplete: it does not consider how the diagnosis is to be used, or the utility associated with either the diagnosis or the treatment of the abnormalities. We argue that the most appropriate definition of (optimal) diagnosis should be based on what the diagnosis is used for. The point of this paper is to question current approaches to formalising diagnostic reasoning, and hopefully focus attention on crucial questions not being studied.

The remainder of the paper is organised as follows. Section 2 introduces the notation and discusses what the diagnostic problem should entail. Section 3 formally defines the six approaches to diagnosis studied in this paper. Section 4 shows examples of how the diagnoses produced by the different approaches are qualitatively different. Section 5 discusses the proposals, evaluating their strengths and weaknesses. Finally, Section 6 draws a few conclusions.

## 2 WHAT IS THE DIAGNOSTIC PROBLEM?

### 2.1 Notation

We call $E$ the knowledge used to compute a diagnosis. $E$ can be broken down into a set $F$ of facts which are unchanging from instance to instance (e.g. $F$ can be a model of a circuit which is being diagnosed), and a set $O$ of observations concerning a particular instance. We call $H = \{h_1, ..., h_m\}$ the set of hypotheses under consideration given $E$. $T = \{t_1, ..., t_l\}$ is the set of possible treatments.

This diagnostic problem can be formalised in either probabilistic or logical terminology. In terms of logical terminology, we use the Theorist framework of hypothetical reasoning [Poole *et al.*, 1987, Poole, 1987], a formalism well suited to the task as the paradigms can be naturally represented in the simple formal framework. Theorist [Poole, 1987] is defined as follows. The user provides $F$, a set of closed formulae (called the *facts*) and $H$, a set of open formulae (called the *possible hypotheses*). A **scenario** is a set $D \cup F$ where $D$ is a conjunction of hypotheses $D = \bigwedge_i h_i$ for some $i = 1, ..., m$, such that $D \cup F$ is consistent. An **explanation** of formula $g$ is a scenario that logically implies $g$. An **extension** is the set of logical consequences of a maximal (with respect to set inclusion) scenario.

For a given treatment $\tau \subseteq T$, we define a utility function $u(E, D, \tau)$. Using a standard decision-theoretic approach (e.g. [Berger, 1985]), the goal of diagnostic reasoning can be defined as choosing $\tau$ to maximise $u(E, D, \tau)$. If there are probability distributions defined over $E$ and $D$, then the maximum expected utility, $\mathcal{E}[u(E, D, \tau)]$, is required. For example, if for diagnosis $i$, $i = 1, ..., k$, the utility associated with treating diagnosis $D_i$ is $u(E, D_i, \tau) = \alpha_i$, and diagnosis $D_i$ occurs with probability $p(D_i)$, the goal is to choose $\tau$ to maximise $\sum_i (\alpha_i \cdot p(D_i))$. In computing expected utility values, influence diagrams [Howard and Matheson, 1981, Shachter, 1986] can be used to find the treatment with highest utility.

### 2.2 Diagnostic Problem

A complete diagnostic cycle consists of reasoning from evidence $E$ to hypothesised diagnosis $H$, and then administering a treatment $T$ for the diagnosis (or abnormalities). This is shown in Figure 1.

In most current formal theories of diagnosis (e.g. [Reiter, 1987, Peng and Reggia, 1987a, de Kleer and Williams, 1987, Pearl, 1987, Pople, 1982]), the treatment phase is not considered at all, and diagnoses are defined with respect to the evidence-hypothesis phase only. These approaches ignore utility considerations totally. It is interesting to note that in one of the earliest analyses of medical diagnostic reasoning, Ledley and

Figure 1: A complete diagnostic cycle

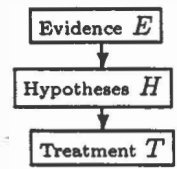

Lusted [1959] described a method of implementing the full diagnostic cycle.

We argue that diagnostic reasoning must take into account the complete cycle, and should consider utility maximisation. In using a utility-based approach, the definition of a diagnosis is strongly influenced by how the diagnosis is to be used.

There are a number of possible uses of a diagnosis, including:

1. to find out the thing (or things) that is wrong; that is, through testing, to determine the exact state of the world with respect to the symptoms observed;

2. to give a plausible account (an explanation) for the symptoms; that is, to give a description of what is wrong that is understandable to people;

3. to enable a decision as to how to fix something; that is, to maximise the utility derived from the diagnostic process through the treatment of the abnormalities;

4. to fix the symptoms; in some cases we may be happy to fix the symptoms without caring about what is really wrong.

The first of these may be carried out by someone who is trying to determine what errors occur so that they can be prevented (for example, by redesigning some components). The second and the third are both activities that an ideal doctor must undertake. As well as giving optimal treatment they also have to be able to give a rationale for the treatment. The fourth may be something that has to be done in an emergency, for example reducing the temperature of a patient with fever, or restoring power in a failed power station.

While each of these may seem reasonable, we will see that different goals will lead to different evaluation criteria. For example, the best decision may involve averaging over many cases, which may not be conducive to a good explanation. Also there may be no point in finding the exact causes for a problem if further refinement of the problem will not improve the outcome. In computing explanations, the most coherent explanation may not contain all the evidence or hypotheses.

It is important to keep these different goals in mind when considering different proposals.

## 3 DESCRIPTION OF PROPOSALS

### 3.1 Most likely single-fault hypothesis

In this approach, the hypotheses considered are the unit hypotheses, $h_1, ..., h_m$. A single-fault diagnosis is defined as a conjunction of hypotheses, only one of which is true. Hence, if hypothesis $h_i$ is the proposed single-fault hypothesis, the single-fault diagnosis $H_i$ is $\overline{h_1} \wedge \overline{h_2} \wedge \cdots \wedge h_i \wedge \cdots \wedge \overline{h_m} =ह_i \wedge \bigwedge_{j \neq i} \overline{h_j}$. For example, if an electrical circuit contains $m$ components, a single-fault diagnosis is that component $i$ is faulty and all other components are functioning normally. The most likely diagnosis is defined as the diagnosis with the highest probability, $p(H_i|E)$, over the set $i \in \{1, ..., m\}$.

## 3.2 Most likely posterior hypothesis

As in the single-fault approach, the hypotheses considered are the unit hypotheses, $h_1, ..., h_m$. The most likely diagnosis is defined as the hypothesis with the highest posterior probability $p(h_i|E)$.

This is the approach taken in MEDAS [Ben-Bassat et al., 1980], INTERNIST [Pople, 1982], and by Pearl [1986], for example. Pearl frames his analysis within Bayesian networks. This entails defining a causal graph of the problem, where the nodes represent random variables (or propositions) and directed edges represent direct causal influences between random variables.

## 3.3 Most likely interpretation

This approach entails considering the set $\mathcal{I} = \{I_1, ..., I_k\}$ of interpretations, the set of truth assignments for the propositions in $H$. The interpretation which is most likely given the evidence must be determined. Pearl [1987] defines this optimal interpretation as the interpretation which has the highest posterior probability given the evidence, $p(I_l|E)$, where there is no $I_j$ such that $p(I_l|E) < p(I_j|E)$. Reiter and Mackworth [1990] advocate considering all visual interpretations for a given image formalised as a set of logical clauses. Note that all interpretations need not be computed in order to find the most likely one [Pearl, 1987].

## 3.4 Probability of provability

In this case we use a logical axiomatisation of the problem, as well as a probabilistic model of the domain. The logical model is used to prove the logical possibilities of the observations $O$ (which constitute the set of hypotheses in which one is interested), and then the probabilistic model is used to compute the likelihood of these. This corresponds to finding the most likely consistency-based diagnosis [Reiter, 1987, Poole, 1989b, de Kleer et al., 1990].

In terms of normal and abnormal function of system components, a consistency-based diagnosis is defined as:

**Definition 3.1 (Consistency-Based Diagnosis)**
A **consistency-based diagnosis** is a minimal set of abnormalities such that the observations are consistent with all other components acting normally [Reiter, 1987].

This approach entails axiomatising the problem as a logical theory $\mathcal{T}$ which consists of a set $\Sigma$ of clauses representing $O$ and $F$. From $\Sigma$ a set $\Gamma$ of clauses logically entailed by $\Sigma$ can be derived. This set can be defined as a minimal disjunct of maximal conjuncts of elements of $H$ that follow from $\Sigma$,[1] i.e.

$$O \wedge F \models (h_i \wedge ... \wedge h_j) \vee (h_k \wedge ... \wedge h_l) \vee ... \vee (h_m \wedge ... \wedge h_n).$$

Each conjunct is defined to be a diagnosis.

In order to compute the probabilities of the elements of $\Gamma$, a probability distribution (or some other measure) must be assigned to $\Sigma$. Then the measure assigned to the $\gamma_i \in \Gamma$ must be computed.

One method of such a computation is provided by the ATMS, as described in [de Kleer and Williams, 1987] or [Provan, 1990, Laskey and Lehner, 1990].[2] An assumption can be assigned to each clause $\Sigma_i \in \Sigma$ to symbolically represent the measure assigned to $\Sigma_i$. The ATMS then computes the set of assumptions assigned to each literal consistent with $\Sigma$. Consequently, the assumption set associated with each $\gamma_i$ can be computed. Assigning a measure to each assumption enables the measure for each $\gamma_i$ to be derived.

The causal relationship in this approach has been described by Pearl [1988] as evidential, as the direction of causality proceeds from evidence to cause.

## 3.5 Covering explanations

In this case the goal is to abduce a causal explanation of the observations.

**Definition 3.2 (Abductive Diagnosis)** Given a causal axiomatisation of the system, an **abductive diagnosis** is a minimal set of hypotheses which, together with the background knowledge implies the observations $O$ [Poole et al., 1987, Poole, 1989a].

As in the provability approach, a logical axiomatisation is required. More formally, the hypotheses considered are the minimal conjunct of elements of $H$ that imply $O$ from $F$ (cf. [Poole, 1987]):

$$F \wedge ((h_i \wedge ... \wedge h_j) \vee (h_k \wedge ... \wedge h_l) \vee ... \vee (h_m \wedge ... \wedge h_n)) \models O.$$

Each conjunct is an explanation or diagnosis.

Abduction is used to compute a causal explanation for the observations. Note that the background knowledge $F$ must be axiomatised differently in this approach and the previous one [Poole, 1988, Poole, 1989a].

Another method of describing this approach using graph theory is that of a causal bipartite graph [Reggia et al., 1985]. In such a graph the bipartite edge sets consist of cause nodes and observation nodes, with directed edges from cause nodes to observation nodes. A node covering of causes for a given set of observation nodes is required.

A probabilistic analysis of this approach can be found in [Peng and Reggia, 1987a, Peng and Reggia, 1987b] and in [Neufeld and Poole, 1987]. Peng and Reggia [1987b] and Neufeld and Poole [1987] describe a method of computing the best diagnoses by evaluating the best partial diagnoses only.

---

[1] See, for example, [Reiter, 1987], [de Kleer and Williams, 1987] or [de Kleer et al., 1990] for details.

[2] For the purposes of this paper it is irrelevant what type of measure or combination function is used. What is of interest here is the measure assigned to $\Gamma$, and what is contained in $\Gamma$.

Pearl [Pearl, 1988] describes this as a causal approach, as the direction of causality proceeds from cause to evidence.

## 3.6 Utility-based explanation

This approach computes not the most probable conjunction of hypotheses (or diagnosis), but the conjunction of hypotheses that are most useful for the treatment phase of diagnosis.

To compute utilities, we can use the definition of a Bayesian network (as done in the first two approaches described in the paper) augmented with a set of decision nodes, a value node and utility values (an influence diagram [Shachter, 1986]).

This approach makes no commitment as to which set of hypotheses constitute a diagnosis, unlike the logic-based approaches (approaches 3 and 4). Bayesian networks can also be used to compute the probability of any conjunction of hypotheses, by creating a new node that represents the conjunction of the hypotheses of the nodes it is influenced by.

Dependent on the utilities of a given problem instance, different combinations of hypotheses will be considered. In general, utility can be assigned on a problem by problem basis. Obviously, dependent on the utility function chosen, this approach can end up computing different combinations of hypotheses than for the "diagnosis" from the first five approaches.

The utility-based approach is influenced by work in computer vision, planning and in cognitive science in which a high-level description of the task influences both the objective function to be evaluated and the method of solving the task. For example, in computer vision, model-based object recognition systems use a description of the object to speed recognition of the object by looking for only image primitives which will most likely constitute a part of the object [Chin and Dyer, 1986].

We argue that the problem representation must be dependent on the nature of the diagnosis required. Thus, if distinguishing among diagnoses does not affect the decision made, there is no point to computing the different diagnoses, or an optimal diagnosis. Or, if the utility value is indifferent to particular liver diseases or heart diseases, the problem can be reformulated (e.g. all liver diseases considered as a "group", and all heart diseases considered as another "group").

As another example, consider a computer which has four main circuit boards, each of which can be divided into groups of components. If the computer goes down due to hardware failure, then the optimal diagnosis for a situation in which the computer must be fixed as soon as possible is to identify which circuit boards need to be replaced. If there is no time pressure, the optimal diagnosis may be to identify which group of components, or which specific components, must be replaced, given the high cost of replacing entire boards.

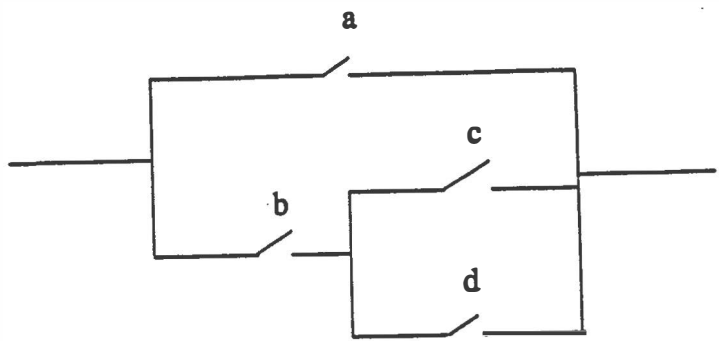

Figure 2: A circuit

## 4 ARE THEY REALLY ALL THE SAME?

In this section we demonstrate that the approaches do indeed give different answers.

**Example 4.1** Consider the analogue circuit of figure 2. In this figure $a$, $b$, $c$ and $d$ are gates that are faulty if they are closed (i.e., they are supposed to be open, but they can be shorted). Let $A$ be the proposition that is true if $a$ is shorted, and $B$ be the proposition that is true if $b$ is shorted, etc.

Suppose we have the knowledge that the gates break independently and we have the priors

$$\begin{aligned} p(A) &= 0.016 \\ p(B) &= 0.1 \\ p(C) &= 0.15 \\ p(D) &= 0.1 \end{aligned}$$

Suppose we observe that there is a current flowing through the circuit. Let $E$ be the proposition that represents this evidence. The diagnoses computed by each of the approaches are now examined. We first give the interpretations as the the probabilities of all interpretations serves as a generator for all the probabilities.

The 16 possible interpretations are:

$$\begin{aligned} p_0 &= p(A \wedge B \wedge C \wedge D|E) & = 0.0006 \\ p_1 &= p(A \wedge B \wedge C \wedge \neg D|E) & = 0.0055 \\ p_2 &= p(A \wedge B \wedge \neg C \wedge D|E) & = 0.0035 \\ p_3 &= p(A \wedge B \wedge \neg C \wedge \neg D|E) & = 0.0313 \\ p_4 &= p(A \wedge \neg B \wedge C \wedge D|E) & = 0.0055 \\ p_5 &= p(A \wedge \neg B \wedge C \wedge \neg D|E) & = 0.0497 \\ p_6 &= p(A \wedge \neg B \wedge \neg C \wedge D|E) & = 0.0313 \\ p_7 &= p(A \wedge \neg B \wedge \neg C \wedge \neg D|E) & = 0.2816 \\ p_8 &= p(\neg A \wedge B \wedge C \wedge D|E) & = 0.0377 \\ p_9 &= p(\neg A \wedge B \wedge C \wedge \neg D|E) & = 0.3395 \\ p_{10} &= p(\neg A \wedge B \wedge \neg C \wedge D|E) & = 0.2138 \\ p_{11} &= p(\neg A \wedge B \wedge \neg C \wedge \neg D|E) & = 0.0 \\ p_{12} &= p(\neg A \wedge \neg B \wedge C \wedge D|E) & = 0.0 \\ p_{13} &= p(\neg A \wedge \neg B \wedge C \wedge \neg D|E) & = 0.0 \\ p_{14} &= p(\neg A \wedge \neg B \wedge \neg C \wedge D|E) & = 0.0 \\ p_{15} &= p(\neg A \wedge \neg B \wedge \neg C \wedge \neg D|E) & = 0.0 \end{aligned}$$

There are a few things to notice about the probabilities of the interpretations:

1. All of the other posterior probabilities can be generated from the $p_i$. In particular, for any formula $w$ we have
$$p(w|E) = \sum_{\{I: w \text{ is true in } I\}} p(I|E)$$

2. All of the interpretations that are not possible have probability zero. The last 5 interpretations are not possible given the circuit, and so must have probability zero. Note that there was no need to do any logical pruning before the probability phase. The logical axiomatisation of the provability and covering cases (4 and 5) was not to remove impossible interpretations (as in [Ledley and Lusted, 1959]) but to determine what it is that we are getting the probability of.

**Single-fault.** The possible single-fault diagnoses are obtained by saying that all of the interpretations except $p_7$, $p_{11}$, $p_{13}$ and $p_{14}$ are impossible. There is only one possible single fault diagnosis: $a$ is shorted. This does not depend on any prior probabilities, except for the knowledge that the probability is zero for all of the impossible diagnoses.

**Posterior.** When computing the most likely posterior, we compare

$$\begin{aligned} p(A|E) &= p_0 + p_1 + p_2 + p_3 + p_4 + p_5 + p_6 + p_7 \\ p(B|E) &= p_0 + p_1 + p_2 + p_3 + p_8 + p_9 + p_{10} + p_{11} \\ p(C|E) &= p_0 + p_1 + p_4 + p_5 + p_8 + p_9 + p_{12} + p_{13} \\ p(D|E) &= p_0 + p_2 + p_4 + p_6 + p_8 + p_{10} + p_{12} + p_{14} \end{aligned}$$

Numerically, the posterior probabilities are:

$$\begin{aligned} p(A|E) &= 0.409 \\ p(B|E) &= 0.632 \\ p(C|E) &= 0.439 \\ p(D|E) &= 0.292 \end{aligned}$$

The most likely faulty component is $b$.

**Interpretation.** The highest probability is $p_9$, which indicates that the most likely interpretation is that $b$ and $c$ are shorted, and $a$ and $d$ are not shorted.[3]

**Provability.** In the fourth case we need to axiomatise the circuit:
$$E \Rightarrow A \vee (B \wedge (C \vee D)),$$
and find the probabilities of the resulting diagnoses:

$$\begin{aligned} p(A|E) &= p_0 + p_1 + p_2 + p_3 + p_4 + p_5 \\ &\quad + p_6 + p_7 \\ p(B \wedge C|E) &= p_0 + p_1 + p_8 + p_9 \\ p(B \wedge D|E) &= p_0 + p_2 + p_8 + p_{10}. \end{aligned}$$

---
[3] Note that this diagnosis just coincidentally corresponds to the most likely posterior hypothesis (in that they both have $b$ broken). If we changed the priors slightly to make $p(C) = 0.12$, the most likely interpretation becomes the one with just $A$ true, but the most likely posterior hypothesis is unchanged.

Numerically, the probabilities are:
$$\begin{aligned} p(A|E) &= 0.409 \\ p(B \wedge C|E) &= 0.383 \\ p(B \wedge D|E) &= 0.256. \end{aligned}$$
The most likely diagnosis is that $a$ is shorted.

**Covering.** For the covering hypothesis case we also need a logical axiomatisation, such as
$$\begin{aligned} A &\Rightarrow E \\ B \wedge C &\Rightarrow E \\ B \wedge D &\Rightarrow E \end{aligned}$$
When $E$ is observed, we get the same comparisons as the previous case, and the same most likely diagnosis.

**Utility-based.** The final case is where we have to take utilities into account. The next three examples show how this can interact with the diagnoses.

**Example 4.2** Suppose that we can fix up the gates of example 4.1 independently. In this case the only relevant probabilities are the posteriors of each of the hypotheses.

We denote by $F_x$ the treatment of fixing gate $x$. Consider the following utility values:
$$u(E, D_x, \tau) = \begin{cases} +\$1 & \text{if } F_x \in \tau \wedge X \in D_x \\ -\$1 & \text{if } F_x \in \tau \wedge X \notin D_x \\ 0 & \text{otherwise} \end{cases}$$

In this case, since $p(B|E)$ is greater than 0.5, we expect to gain from fixing gate $b$, but do not gain from fixing any of the other gates. Thus it is only worthwhile fixing gate $b$.

If one believes that the aim of the diagnosis is to fix all of the problems, then this is a peculiar thing to do. Based on the logical analysis, it cannot be the case that only $b$ is shorted.

**Example 4.3** Suppose there is a heavy penalty for not fixing the circuit by replacing a particular gate, as shown in the following utility function:
$$u(E, D_x, T_{D_x}) = \begin{cases} +\$1 & \text{if } F_x \in \tau \wedge X \in D_x \\ -\$1 & \text{if } F_x \in \tau \wedge X \notin D_x \\ -\$10 & \text{if } F_x \notin \tau \wedge X \in D_x \\ 0 & \text{if } F_x \notin \tau \wedge X \notin D_x \end{cases}$$

In this case, all gates will be replaced in our example. All we needed to compute was the posterior probabilities for the individual hypotheses.

**Example 4.4** Suppose we can fix up gates $b$ and $c$ independently, but that the ways to fix up gates $a$ and $d$ interact in a complex way. In this case the relevant probabilities are

$$\begin{aligned} p(A|E) &= 0.409 \\ p(B|E) &= 0.632 \\ p(C|E) &= 0.439 \\ p(A \wedge D|E) &= 0.041 \\ p(A \wedge \neg D|E) &= 0.368 \\ p(\neg A \wedge D|E) &= 0.251 \\ p(\neg A \wedge \neg D|E) &= 0.340 \end{aligned}$$

which is not what any of the diagnoses computed. Note that in this case the probabilities that are needed are not determined by what can explain the observations, but rather what is needed for the treatment.

It might coincidentally happen that two approaches compute the best qualitative diagnosis, but this is not true in general. It is, however, not coincidence that the provability and covering approaches produce the same answer. It can be shown that, under certain reasonable assumptions about how the knowledge is represented, the propositional versions of the abductive and the deductive systems [Poole, 1988] are identical. This is not, however, necessarily true for the non-propositional versions. The difference arises because of the level of detail of the diagnoses. The more specific the diagnoses, the less likely it is. The abductive systems require the level of detail specific enough to imply the observations. In the deductive system, the level of detail is prescribed, not by the observation, but by the knowledge base [Poole, 1989a].

## 5 ANALYSIS OF PROPOSALS

Each of the proposals has good and bad points, some of which are now discussed.

### 5.1 Single-fault Diagnosis

The main problem with the most likely single fault hypothesis is that it may be wrong. The real diagnosis may be a combination of faults.

Another problem is that the single fault definition depends on the representation used. At one level of abstraction a single fault may be a very complicated combination of faults at another level of detail. For example, at one level of abstraction a problem may be that one power supply is broken. At another level of detail there may be a number of problems with the one power supply.

### 5.2 Most Likely Posterior

There are a number of problems with the "most likely posterior hypothesis" approach:

1. It is the weakest statement about the world. Thus, if a pigeon is a type of bird, one must have $p(pigeon|E) \leq p(bird|E)$.

2. High (possibly irrelevant) priors may dominate the relevant hypotheses.

3. Groups of hypotheses are not considered. This approach cannot compute diagnoses which consist of sets of hypotheses. Multiple-hypothesis diagnoses can be determined by heuristics only (e.g. as is done in INTERNIST [Pople, 1982]). In contrast, the interpretation, consistency and covering methods can compute multiple-hypothesis diagnoses based on the underlying theory of the method. Note that this does not imply we cannot diagnose systems with multiple faults, but rather that we do not consider conjunctions in diagnoses.

Note that we can add hypotheses which are conjunctions of hypotheses, but these will always be less likely that the original hypotheses.

### 5.3 Most Likely Interpretation

In the most likely interpretation approach (# 3) there can be an exponential number of interpretations, and so for any reasonable sized problem we do not want to derive all interpretations (as in [Reiter and Mackworth, 1990]). Many interpretations will be highly unlikely, and computing them will be a waste of computational resources. However, it is not necessary to enumerate all interpretations [Pearl, 1987, de Kleer and Williams, 1989][4].

Another drawback of this approach is that the most likely interpretation does not let us be agnostic about any proposition; we have to give a truth value to every hypothesis, no matter how related to the observations. Changing the space of hypotheses can change the probability of the most likely composite, and even what the most likely composite is [Pearl, 1988, p.285]. Even more importantly, for building large knowledge bases adding "irrelevant" hypotheses can also change the most likely interpretation. If we imagine a random variable that is probabilistically independent of the other variables, then the most likely qualitative conjunct will remain the same: we will end up with the product of the most likely prior of the irrelevant hypothesis and the most likely interpretation. If however the new random variable is not independent, then just by imagining different scenarios we can change the qualitative diagnosis, as the following example shows.

**Example 5.1** Consider a patient who has symptoms which suggest either the flu or yellow fever, the flu being the more likely diagnosis. Now, suppose we add a variable that represents where the person was at some particular time two weeks previously. We partition the world so that the different places all have equal probabilities (e.g., "Africa" may be one area, and "above the most northerly floor tile in their office" may be another). If these new variables are independent of the disease variable, the most likely interpretation will consist of the most likely values for each variable. Simply by assuming that it is much more likely that the patient has yellow fever if she was in Africa, the most likely composite hypothesis may be that the patient has Yellow fever and was in Africa[5].

One other problem with finding the most likely interpretation is that the most likely interpretation may have very low probability. Peng and Reggia [1987c] first pointed this out and suggested that rather than finding the most likely interpretation, we should find a set of interpretations that covers some percentage of all of the cases.

---

[4]Note that the phase of removing interpretations inconsistent with the knowledge (i.e., those interpretations that are not models of the logical constraints) is unnecessary because $p(h_i|E) = 0$ if $h_i$ is inconsistent with evidence $E$.

[5]This example is different from the example given by Pearl [1988, p. 285], in that we are adding a new hypothesis rather than changing a hypothesis. Just by imagining a new hypothesis, and making no new observations, we can change the value of the most likely interpretation.

### 5.4 Covering Explanations

Approaches 4 and 5 are discussed together because propositional versions of these approaches have been shown to produce the same results under certain assumptions [Poole, 1988]. These differ from the interpretation approach in enabling us to be agnostic about the value of some variables which are irrelevant to the goal.

### 5.5 Consistency and Covering Approaches

These logic-based approaches are limited to what is provable within the logical theory $\mathcal{T}$. Hence, they are sensitive to the particular logical description. If an observation is not provable, then nothing can be said about it. (Similarly, for the Bayesian approach, if the appropriate conditional probabilities are present for an observation, no measure can be assigned to the observation.) In contrast, in the utility-based approach considered here, what is computed depends on the goal of the diagnosis.

Both the consistency and covering approaches rely on making assumptions about the system under examination. Pearl [1990] has shown that this approach has two major flaws: (1) independencies among events with unknown probabilities cannot be represented; and (2) domain knowledge describing defeasible conditional sentences cannot be represented. Thus, these approaches are limited to problems in which the domain knowledge can be defined in categorical terms, e.g. strict taxonomic hierarchies, deterministic systems (electronic circuits but not medical diagnosis or default reasoning), etc. If these approaches are misapplied to certain domains, non-intuitive results can be obtained.

Pearl [1990] argues that instead of making assumptions about the hypotheses, examining the logical consequences of these assumptions, and then assigning a probability measure over the assumptions, a probability measure should be assigned over the logical clauses $\Sigma$ and the probabilities assigned to interpretations be examined (as done in approach 2). If not all clauses are assigned a weight, then probability bounds (specifically the inner and outer measure) can be obtained for the measure assigned to the hypotheses. For example, Nilsson [1986] presents a technique which can accept an incomplete probabilistic specification, and computes ranges for the required measures for hypotheses. This paper argues that it is not clear that the approach suggested by Pearl (i.e. that the most likely interpretation must be evaluated) is always the best one.

However, because these logic-based approaches seem to be appealing for a number of reasons, it may be important to determine subcases of non-categorical domains in which paradoxes do not arise.

### 5.6 Utility-based diagnoses

The utility-based approach assumes that utilities are domain-dependent, which implies that diagnoses must be domain-dependent as well. In contrast, the other approaches assume that the definition of diagnosis is domain-independent. In addition, preferences used to define best diagnoses cannot be assigned in a domain-independent manner. Doyle and Wellman [1989] have proven that there exists no universally-valid preference set; in other words, preferences are consistent only within particular domains. Furthermore, Doyle [1989] argues for a decision-theoretic, and not merely probabilistic, definition of consistency and rationality in decision-making (and therefore diagnosis).

The utility-based approach is considered the proper one in decision theory and in several areas of cognitive science. For example, in visual recognition it has been shown how several low-level (and quite primitive) pre-attentive processes are used to focus attention on the most salient features of a scene [Treisman, 1982], and simple features are used to guide scene interpretation [Rosenfeld, 1987]. In some cases, results from these pre-attentive processes are used directly to initiate action. For example, in the forest if a deer sees motion close by (the presence of motion in the visual field can be detected pre-attentively), it will start running without identifying the cause of the motion, as it might get eaten if it spent the time trying to identify the cause of the motion. If the deer is not in danger, motion can trigger a closer scrutiny for the cause of the motion.

One of the problems of the utility-based approach is that there is nothing in a diagnosis unless there is a goal and utilities. There is no "value free" definition of a diagnosis. Whether this is desirable or not is left up to the reader. There is also the problem of being able to give someone an understandable explanation to answer the question "what is wrong?".

## 6 CONCLUSIONS

This paper has analysed several definitions of what a diagnosis, and so, presumably, what a most-likely diagnosis is, showing them to be mutually incompatible. It is not clear that one definition is correct over all domains and situations, and for all possible uses that there could be for the diagnosis. Instead, we argue that the notion of most likely diagnosis cannot be defined *a priori*, but is defined based on what the diagnosis is to be used for, and on the uitlity of the outcomes of treating the abnormalities. In other words, there may be no *a priori* ontological definition of optimal diagnosis; it is epistemological and situation-dependent.